\documentclass[journal,twoside,web]{ieeecolor}
\usepackage{generic}
\usepackage{cite}
\usepackage{amsmath,amssymb,amsfonts}
\usepackage{graphicx}
\usepackage{hyperref}
\hypersetup{hidelinks=true}
\usepackage{textcomp}
\usepackage{booktabs}
\usepackage{multirow}
\usepackage{makecell}
\usepackage{url}
\providecommand{\refname}{References}

\def\BibTeX{{\rm B\kern-.05em{\sc i\kern-.025em b}\kern-.08em
    T\kern-.1667em\lower.7ex\hbox{E}\kern-.125emX}}

\begin{document}

\title{EyeMVP: OCT-Informed Fundus Representation Learning via Paired CFP--OCT Pretraining}

\author{Zhuo Deng\textsuperscript{*},
        Ruiheng Zhang\textsuperscript{*},
        Ziheng Zhang,
        Weihao Gao,
        Yitong Li,
        Qian Wang,
        Lei Shao,
        Jiaoyue Dong,
        Zhixi Zeng,
        Lijian Fang,
        Haibo Wang,
        Xiaobin Lin,
        Tao Liu,
        Zhicheng Du,
        Zhengwei Zhang,
        Lin Yang,
        Zheng Gong,
        Xinyu Zhao,
        Zhenquan Wu,
        Fang Li,
        Zhiguang Zhou,
        Guoming Zhang,
        Sun Jing,
        Han Lv,
        Wenbin Wei\textsuperscript{\dag},
        \IEEEmembership{and}
        Lan Ma\textsuperscript{\dag}
\thanks{$^{*}$Zhuo Deng and Ruiheng Zhang contributed equally to this work.}
\thanks{$^{\dag}$Corresponding authors: Wenbin Wei (weiwenbintr@163.com) and Lan Ma (malan@sz.tsinghua.edu.cn).}
\thanks{Z. Deng, Z.H. Zhang, W.H. Gao, Z.C. Du, Z. Gong, F. Li, and L. Ma are with Shenzhen International Graduate School, Tsinghua University, Shenzhen, China.}
\thanks{R. Zhang, Y. Li, Q. Wang, L. Shao, J. Dong, Z.X. Zeng, and W. Wei are with Beijing Tongren Eye Center, Beijing Tongren Hospital, Capital Medical University, Beijing, China.}
\thanks{L. Fang is with Liangxiang Hospital of Beijing Fangshan District, Capital Medical University, Beijing, China.}
\thanks{H. Wang is with The Third People's Hospital of Dalian, Dalian, China.}
\thanks{X. Lin, L. Yang, and Z.G. Zhou are with the National Clinical Research Center for Endocrine and Metabolic Diseases, The Second Xiangya Hospital of Central South University, Changsha, China.}
\thanks{T. Liu is with The Central Hospital of Baoji City, Baoji, China.}
\thanks{Z.W. Zhang is with Wuxi No.2 People's Hospital, Affiliated Wuxi Clinical College of Nantong University, Wuxi, China.}
\thanks{X. Zhao, Z.Q. Wu, and G. Zhang are with Shenzhen Eye Hospital, Southern Medical University, Shenzhen, China.}
\thanks{S. Jing and H. Lv are with Beijing Friendship Hospital, Capital Medical University, Beijing, China.}
\thanks{This work was supported by Development and Reform Commission of Shenzhen Municipality (F-2024-Z99-503929), Shenzhen Medical Research Fund (C2401007), National Natural Science Foundation of China (82220108017, 82141128), The Capital Health Research and Development of Special (2024-1-2052), Science \& Technology Project of Beijing Municipal Science \& Technology Commission (Z201100005520045), and Sanming Project of Medicine in Shenzhen (SZSM202311018).}
}

\maketitle

\begin{abstract}
Color fundus photography (CFP) is the mainstay for large-scale retinal
screening, yet its diagnostic capacity is constrained by the lack of
depth-resolved structural information. Optical coherence tomography (OCT)
provides cross-sectional retinal anatomy, but is less accessible in
population-level screening. Here, we present EyeMVP, a cross-modal retinal
foundation model that uses paired CFP--OCT pretraining to learn
OCT-informed CFP representations. EyeMVP is
pretrained on 674,893 strict same-eye same-day paired CFP--OCT image triples from 112,642 patients
across eight hospitals in China. The model
uses cross-modal masked reconstruction to enrich CFP representations with
OCT-associated supervision, while requiring only CFP images at inference. To
accommodate the non-aligned imaging geometry between en-face CFP and
cross-sectional OCT, EyeMVP combines source-constrained cross-attention
with CFP-derived structural masks. Across
15 downstream tasks, including classification, segmentation, and
few-shot adaptation, EyeMVP performs on par with or better than representative
retinal foundation models overall, with consistent gains on tasks involving
macular and optic nerve structure. For CFP-challenging macular diseases,
EyeMVP achieves an AUROC of 0.923 for macular edema and 0.867 for myopic
macular schisis. Qualitative t-SNE visualization of the frozen encoder embeddings further
shows that EyeMVP produces more class-coherent CFP representations than
competing foundation models. In an exploratory reader study, EyeMVP
exceeds junior and intermediate ophthalmologist groups but does not reach
senior ophthalmologist performance on macular edema, while showing
numerically higher balanced accuracy than all reader groups on myopic macular
schisis. These results suggest that pixel-level cross-modal reconstruction
can enrich CFP representations with OCT-associated supervision,
providing a practical route toward stronger CFP-based retinal analysis in
screening settings.
\end{abstract}


\section{Introduction}
\label{sec:introduction}

\IEEEPARstart{C}{olor} fundus photography (CFP) is the mainstay of
large-scale retinal screening because it is inexpensive, rapid, and widely
available. However, CFP is a two-dimensional en-face projection of the
retinal surface and therefore provides only indirect information about
depth-resolved retinal structure. Many clinically important abnormalities,
including macular edema, myopic macular schisis, intraretinal fluid, and
optic nerve head deformation, are defined or confirmed by cross-sectional
anatomy rather than by surface appearance alone. Optical coherence tomography
(OCT) provides this layer-resolved structural information and has become
central to modern retinal diagnosis, but its availability in population-level
screening and resource-limited settings remains more restricted than CFP
\cite{bourne2021trends,mohammadpour2020diagnostics}. This gap motivates a practical
representation learning question: can paired CFP--OCT data be used during
pretraining to enrich CFP representations with OCT-associated supervision,
while requiring only CFP images at inference?

Recent retinal foundation models have shown that self-supervised pretraining
can substantially improve the transferability of ophthalmic image
representations \cite{ericsson2022self,le2020contrastive,deng2024ophglm}. Unimodal masked autoencoders such as RETFound
\cite{zhou2023foundation} learn strong CFP features from large-scale fundus
collections and generalize across multiple downstream tasks. Nevertheless,
models trained only on CFP are limited by the information available in the
input modality. They can learn rich appearance and texture patterns, but they
receive no direct learning signal from cross-sectional retinal anatomy. This
is particularly limiting for CFP-challenging diseases whose visual signs are
subtle, indirect, or partially occult in fundus photographs.

A natural strategy is to incorporate OCT during pretraining. Existing
multimodal ophthalmic models, including VisionFM \cite{qiu2024development},
EyeCLIP \cite{shi2025multimodal}, Eyefound \cite{shi2024eyefound}, and
MIRAGE \cite{morano2025multimodal}, have explored multimodal representation
learning across retinal imaging modalities. Many of these approaches rely on
image-level contrastive alignment or related global objectives, following the
broader success of contrastive image-language pretraining \cite{radford2021learning}. Such
objectives are effective for learning semantic correspondence between
modalities, but they do not explicitly exploit the pixel- or patch-level
structural information available in paired CFP--OCT data. For diseases whose
diagnostic evidence depends on localized macular or optic nerve structure,
global alignment may underuse the most informative cross-modal supervision.

Masked reconstruction provides a more direct way to use cross-modal
supervision between modalities. In principle, reconstructing OCT from CFP
tokens can encourage the model to encode CFP features that are predictive of
underlying retinal structure. However, applying standard multimodal masked
autoencoding to CFP--OCT pairs is not straightforward. Frameworks such as
MAE \cite{he2022masked} and MultiMAE \cite{bachmann2022multimae} assume that modalities share a common
spatial grid, as in RGB-depth or RGB-segmentation pairs. CFP and OCT violate
this assumption: CFP represents the retinal surface in an en-face coordinate
system, whereas OCT samples cross-sectional depth structure. A CFP patch and
an OCT patch with the same image index do not correspond to the same physical
retinal location. This non-aligned imaging geometry makes naive
patch-level cross-modal reconstruction prone to learning superficial
correlations rather than useful structural correspondence.

In this work, we present EyeMVP, a cross-modal retinal foundation
model designed to learn OCT-informed CFP representations from strict same-eye
same-day paired CFP--OCT images. EyeMVP uses cross-modal masked
reconstruction as the pretraining signal, but adapts it to the non-aligned
geometry of CFP and OCT through two design choices. First, a source-constrained cross-attention
decoder restricts cross-modal reconstruction to source-modality features,
creating an information bottleneck that discourages local texture shortcuts.
Second, CFP-derived structural masks provide anatomical anchors, including
the optic disc, macular region, and major vessels, to ground cross-modal
learning without requiring explicit patch-level CFP--OCT registration.

We pretrain EyeMVP on 674,893 strict same-eye same-day paired CFP--OCT image triples from
112,642 patients across eight hospitals in
China. The pretrained CFP encoder is evaluated on 15 downstream tasks,
including retinal disease classification, dense segmentation, and
few-shot adaptation. Across these tasks, EyeMVP is competitive with or improves
over representative retinal foundation models, with the largest gains on
macular and optic nerve tasks where structural information is clinically
important. For CFP-challenging macular disease classification, EyeMVP
achieves an AUROC of 0.923 for macular edema and 0.867 for myopic macular
schisis. An exploratory reader study further suggests that the learned CFP
representation may provide useful support for macular conditions that are
difficult to assess from fundus photographs alone.

The main contributions of this work are:
\begin{itemize}
  \item We propose EyeMVP, an OCT-informed CFP representation learning
        framework that uses paired CFP--OCT data during pretraining while
        requiring only CFP images at downstream inference.

  \item We introduce a cross-modal masked reconstruction strategy tailored to
        the non-aligned imaging geometry of CFP and OCT, using
        source-constrained cross-attention to encourage prediction from the
        source modality rather than target-modality shortcuts.

  \item We incorporate CFP-derived anatomical masks as structural layout
        guidance during pretraining, providing coarse fundus landmarks without
        requiring explicit patch-level CFP--OCT registration.

  \item We pretrain EyeMVP on 674,893 strict same-eye same-day CFP--OCT image
        triples from 112,642 patients and evaluate it across 15 downstream
        settings, showing consistent gains over representative retinal
        foundation models, especially on macular and optic nerve tasks.
\end{itemize}

\section{Related Work}
\label{sec:related}

\subsection{OCT-Enhanced Fundus Image Analysis}

CFP and OCT provide complementary views of retinal disease. CFP is widely
available and captures en-face retinal appearance, whereas OCT provides
cross-sectional structural information that is often required to confirm or
quantify macular and optic nerve abnormalities \cite{mohammadpour2020diagnostics,wong2019war}. A line of prior work has
therefore investigated whether OCT-derived labels or measurements can improve
fundus-based analysis. Varadarajan et al.~\cite{varadarajan2018predicting}
trained deep learning models to predict OCT-derived center-involved diabetic
macular edema and retinal fluid from fundus photographs, showing that CFP can
contain indirect visual cues associated with OCT-defined disease. Similarly,
Medeiros et al.~\cite{medeiros2018machine} used OCT retinal nerve fiber layer
measurements to supervise a model that predicts glaucomatous structural
damage from optic disc photographs. These studies established the feasibility
of learning OCT-associated information from CFP, but they were primarily
task-specific supervised models rather than general-purpose pretraining
frameworks.

More recent work has explored cross-modal knowledge transfer between fundus
and OCT images. MultiEYE and OCT-CoDA~\cite{wang2024multieye} formulate an
OCT-enhanced disease recognition setting in which OCT information is used
during training to improve fundus-only inference, including through
concept-based distillation from OCT to CFP. This direction is closely aligned
with the practical motivation of our work: OCT should enrich fundus
representations during model development, but downstream deployment should
remain compatible with CFP-only screening. EyeMVP differs in two respects.
First, it learns from strict same-eye same-day paired CFP--OCT images through pixel-level
cross-modal masked reconstruction rather than task-specific disease
distillation. Second, it aims to produce a reusable CFP encoder for
classification, segmentation, and few-shot adaptation, rather than
a model specialized to one disease recognition benchmark.

\subsection{Retinal Foundation Models}

Foundation models have recently become an important direction in ophthalmic
image analysis. RETFound~\cite{zhou2023foundation} demonstrated that masked
autoencoding on large-scale retinal images can produce transferable
representations for disease detection and segmentation. VisionFM
\cite{qiu2024development} further scaled ophthalmic pretraining across
multiple image modalities, tasks, disease categories, devices, and
demographic groups, supporting a generalist ophthalmic AI setting.
Vision-language models such as FLAIR~\cite{silva2025foundation} and EyeCLIP
\cite{shi2025multimodal} incorporate textual supervision or multimodal
contrastive learning to improve zero-shot, few-shot, and long-tail
disease performance. Eyefound~\cite{shi2024eyefound} and
MIRAGE~\cite{morano2025multimodal} also expand foundation modeling toward
broader ophthalmic or retinal modality coverage.

Despite this progress, most existing retinal foundation models are optimized
for either unimodal representation learning or global multimodal alignment.
Unimodal CFP models can learn strong appearance features but do not receive
direct supervision from cross-sectional retinal structure. Multimodal
contrastive or vision-language models improve semantic alignment across
images and text, but their objectives usually operate at the image or report
level and therefore do not explicitly exploit the local structural
correspondence available in paired CFP--OCT data. EyeMVP is complementary to
these efforts: it focuses on using paired OCT as a pretraining signal to
enrich CFP representations with structural information while retaining
CFP-only inference.

\subsection{Multimodal Masked Reconstruction Pretraining}

Masked autoencoders have provided a simple and scalable framework for
self-supervised visual representation learning \cite{ericsson2022self}. MAE~\cite{he2022masked}
trains a vision transformer to reconstruct randomly masked image patches and
has been widely adopted in medical imaging because reconstruction encourages
the model to capture fine-grained spatial structure. MultiMAE
\cite{bachmann2022multimae} extends this idea to multimodal and multitask
settings by reconstructing multiple outputs, such as RGB, depth, and semantic
segmentation, from partially visible modality tokens. Related cross-modal
masked reconstruction methods have also been explored in broader multimodal
pretraining settings \cite{guo2024crossmae}, and in medical
vision-language pretraining; for example, M$^3$AE
\cite{chen2022multi} reconstructs masked image and text tokens to learn
cross-modal medical representations.

However, existing multimodal masked autoencoding frameworks are most natural
when modalities share a common spatial coordinate system or can be treated as
aligned token sequences. This assumption holds for many RGB-depth-segmentation
settings, but it is violated by CFP--OCT pairs. CFP is an en-face projection
of the retinal surface, whereas OCT samples cross-sectional depth structure;
therefore, patch indices across the two modalities do not denote the same
physical retinal location. Naively applying multimodal masked reconstruction
to CFP--OCT can encourage shortcut correlations or unstable cross-modal
optimization rather than robust structural representation learning. EyeMVP
adapts masked reconstruction to this asymmetric imaging geometry through
source-constrained cross-attention and CFP-derived structural anchors.

\section{Methods}
\label{sec:methods}
\subsection{Pretraining Dataset Construction}
\label{sec:data}

Raw multimodal retinal imaging data were collected from eight tertiary
hospitals across China between August
2015 and December 2025. The archive contained 3,058,942 images from 112,642
participants (59,187 females and 53,455 males; mean age 57.5 years, SD 14.6,
range 1--95; all of Chinese ethnicity). CFP images were acquired using five
platforms, including Zeiss CLARUS 500, Canon CR-2, Topcon TRC, SYSEYE, and
Airdoc systems. OCT volumes were acquired using Zeiss CIRRUS, Heidelberg
Spectralis, and SVision systems. Site-level ethics approval was coordinated by
the lead center, Beijing Tongren Hospital, and all data were fully
de-identified before analysis.

CFP examinations included both optic-disc-centered and macula-centered fundus
photographs. OCT examinations included both optic nerve head and macular scan
protocols, reflecting routine clinical imaging for glaucoma and macular
disease assessment. CFP and OCT examinations were linked using patient
identifier, eye laterality, and acquisition date. A valid pair required CFP
and OCT to be acquired from the same eye on the same calendar day and to pass
metadata and image-quality checks. We refer to this criterion as strict
same-eye same-day pairing. Eye laterality was determined from DICOM metadata
and checked against fundus image orientation. For each retained patient-eye
visit, representative OCT B-scans were sampled from the corresponding OCT
volume and linked to the CFP image from the same visit. CFP-derived structural
masks were used to provide coarse fundus-layout information during
pretraining.

A four-stage quality control pipeline was applied. First, each raw image was
screened using an automated image quality assessment algorithm
\cite{gong2025acquire}; images with severe blur, motion artifacts,
under- or over-exposure, pupil occlusion, incomplete retinal field, or poor
OCT signal were removed. Second, same-day same-eye CFP--OCT examinations were
identified from DICOM timestamps and laterality tags. Third, candidate pairs
with missing or inconsistent metadata, implausible acquisition timestamps,
insufficient retinal coverage, or laterality mismatch were excluded. Fourth,
patients included in private downstream evaluation datasets were excluded
from the pretraining pool to reduce data leakage risk.

For each eligible patient-eye visit, CFP images were paired with
representative B-scans from the corresponding OCT volume. Rather than using
only a single central B-scan, we sampled 3--4 representative B-scans from
each retained OCT volume to provide limited spatial coverage of the acquired
scan region. This produced 189,801 eligible patient-eyes and 674,893 final
CFP--OCT image triples, with a mean of 3.56 OCT frames per patient-eye. Each
triple consisted of a CFP image, a paired OCT B-scan, and a CFP-derived
structural mask. A random 5\% subset of the constructed triples was manually
reviewed by clinicians to verify same-eye pairing, image quality, and
segmentation plausibility.

All CFP and OCT images were resized to $512\times512$. OCT B-scans were
represented as three-channel grayscale images to use the same reconstruction
interface as CFP. CFP-derived structural masks were resized to
$256\times256$ and encoded as discrete anatomical labels. A site-level
breakdown of the pretraining data and imaging equipment is provided in
Table~\ref{tab:pretrain_stats}.

\begin{figure*}[t]
  \centering
  \includegraphics[width=\textwidth]{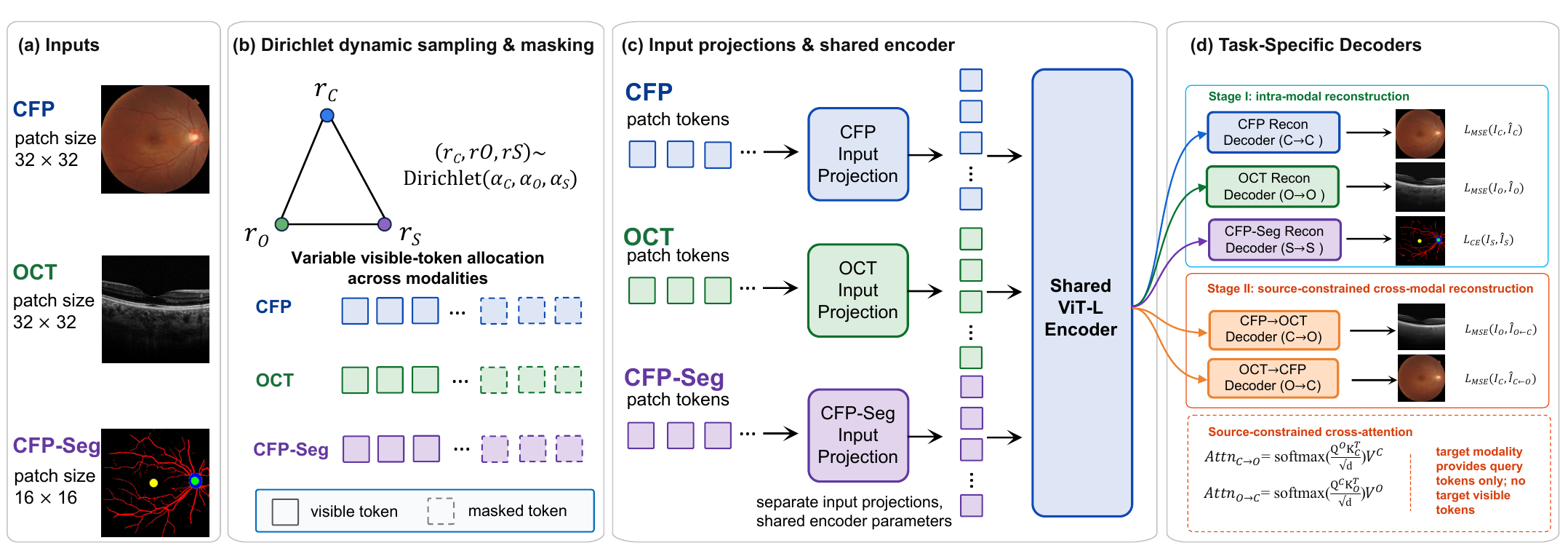}
  \caption{Overview of the EyeMVP pretraining framework. Each pretraining sample
contains a CFP image, a paired OCT B-scan, and a CFP-derived structural mask
(CFP-Seg). Dirichlet sampling dynamically assigns visible token ratios across
modalities, and modality-specific input projections map the visible tokens
into a shared ViT encoder. Stage I uses three intra-modal decoders to
reconstruct CFP, OCT, and CFP-Seg targets. Stage II introduces two
source-constrained cross-modal decoders for CFP-to-OCT and OCT-to-CFP
reconstruction. CFP and OCT reconstruction losses use mean-squared error,
whereas CFP-Seg reconstruction uses cross-entropy.}
\label{fig:workflow}
\end{figure*}

\subsection{CFP-Derived Structural Masks}
\label{sec:cfpseg}

CFP-derived structural masks (CFP-Seg) were used as an auxiliary input during
pretraining. Each mask contains four classes: background, optic disc, foveal
region, and major retinal vessels. The masks were generated offline from CFP
images using task-specific segmentation networks, including U-Net
\cite{ronneberger2015u} for optic disc and foveal region segmentation, and
UNet++ or Attention U-Net \cite{zhou2019unet++, oktay2018attention, deng2024ftsegnet} for
retinal vessel segmentation. These networks were trained on public fundus
segmentation datasets \cite{deng2025fundus} and fine-tuned on held-out in-house annotations that
were not included in pretraining or downstream evaluation. The segmentation
outputs were binarized and merged into a single four-class mask for each CFP
image.

CFP-Seg does not provide disease labels. Its role is to provide coarse
anatomical layout information from the CFP image itself. Because CFP and OCT
are acquired in different imaging geometries, CFP-Seg supplies explicit
fundus landmarks that help condition cross-modal reconstruction on the
observed CFP field.

In this setting, CFP-Seg helps identify the optic disc, foveal region, and
major vessel trajectories in the CFP image, providing structural context about
the current fundus field and spatial layout. CFP-Seg is used only during
pretraining and is discarded during downstream inference. Its contribution is
evaluated by removing this auxiliary input in ablation experiments.

\begin{table*}[t]
\caption{Per-center breakdown of the pretraining dataset.
Cohort demographics and disease distribution are given in
Supplementary Tables~S1 and~S2.}
\label{tab:pretrain_stats}
\centering
\renewcommand{\arraystretch}{1.05}
\setlength{\tabcolsep}{3pt}
\footnotesize
\begin{tabular}{lrll}
\toprule
Site & Patients & CFP & OCT \\
\midrule
Beijing Tongren Hospital        & 57,830 & CLARUS/CR-2/SYSEYE/Airdoc & Spectralis/SVision \\
Fangshan District Hospital      & 26,239 & CR-2                       & Spectralis/SVision \\
Dalian No.3 Hospital            &  9,826 & CR-2                       & Spectralis         \\
Baoji Central Hospital          &  5,837 & CR-2                       & Spectralis         \\
Wuxi No.2 Hospital              &  5,408 & TRC/CR-2                   & CIRRUS/Spectralis  \\
Shenzhen Eye Hospital           &  3,647 & CR-2/TRC                   & Spectralis         \\
Second XiangYa Hospital         &  2,653 & TRC/CR-2                   & CIRRUS/Spectralis  \\
Beijing Friendship Hospital     &  1,202 & CR-2                       & Spectralis         \\
\midrule
\textbf{Total}                  & \textbf{112,642} & 5 platforms & 3 platforms \\
\bottomrule
\end{tabular}
\end{table*}

\subsection{EyeMVP Architecture}
\label{sec:arch}

EyeMVP uses three input modalities during pretraining: CFP, OCT, and CFP-Seg.
Each modality is divided into non-overlapping patches and projected into token
embeddings by a modality-specific patch embedding layer. CFP and OCT inputs
use patch size $32\times32$, and CFP-Seg uses patch size $16\times16$. The
token sequences are then processed by a shared ViT-L encoder
\cite{dosovitskiy2020image}. Thus, different
modalities have separate input projections but share the same encoder
parameters.

The framework contains five lightweight decoders. Three are intra-modal
reconstruction decoders for CFP, OCT, and CFP-Seg:
\[
C \rightarrow C,\quad O \rightarrow O,\quad S \rightarrow S .
\]
These decoders reconstruct masked patches of their own modality from encoded
token features and are used throughout pretraining.

The remaining two decoders are cross-modal reconstruction decoders:
\[
C \rightarrow O,\quad O \rightarrow C .
\]
They are introduced only in the second training stage. The CFP-to-OCT decoder
uses CFP encoded tokens to reconstruct OCT patches, and the OCT-to-CFP decoder
uses OCT encoded tokens to reconstruct CFP patches. These decoders provide an
explicit cross-modal prediction objective, which is not guaranteed by
intra-modal reconstruction alone.

\subsection{Dirichlet Masking Strategy}
\label{sec:masking}

During pretraining, we use a Dirichlet masking strategy to sample visible
token allocation across modalities. For each training sample, modality-wise
visible token proportions are drawn from a Dirichlet distribution:
\begin{equation}
(r_C,r_O,r_S) \sim \mathrm{Dirichlet}(\alpha_C,\alpha_O,\alpha_S),
\end{equation}
where $r_C$, $r_O$, and $r_S$ denote the visible token proportions for CFP,
OCT, and CFP-Seg. Given the total visible token budget, visible tokens are
sampled within each modality according to these proportions, and the
remaining tokens are masked.

This strategy exposes the shared encoder to variable modality availability.
Compared with using a fixed masking ratio for each modality, Dirichlet masking
provides a broader range of reconstruction conditions and reduces dependence
on a single fixed modality configuration. During Stage II, the sampling is
constrained to retain sufficient visible tokens from the source modality so
that CFP-to-OCT and OCT-to-CFP reconstruction receive stable source
information.

\subsection{Source-Constrained Cross-Modal Decoding}
\label{sec:scca}

Intra-modal reconstruction alone does not guarantee that the shared encoder
learns useful CFP--OCT correspondence. When CFP is reconstructed from CFP
features and OCT from OCT features, the model can solve the pretext task
largely through within-modality cues. For high-capacity encoders and
lightweight decoders, local texture and modality-specific appearance
statistics can often support good reconstruction. Thus, a shared encoder with
multiple intra-modal heads may encourage feature sharing, but it does not
explicitly require CFP features to predict OCT structure.

To encourage cross-modal interaction, EyeMVP introduces CFP-to-OCT and
OCT-to-CFP reconstruction in Stage II. A naive cross-modal decoder may still
exploit shortcuts if target-modality visible tokens or mixed multimodal tokens
are available during decoding. In the CFP--OCT setting, such shortcuts are
undesirable because the modalities are not patch-wise registered; local
texture correlations may reduce reconstruction loss without learning useful
structural correspondence.

We therefore use source-constrained cross-attention. For CFP-to-OCT
reconstruction, OCT positional tokens are used as Queries, while CFP encoded
tokens are used as Keys and Values:
\begin{equation}
\mathrm{Attn}_{C\rightarrow O}
=
\mathrm{softmax}
\left(
\frac{Q_O K_C^\top}{\sqrt{d}}
\right)
V_C .
\end{equation}
For OCT-to-CFP reconstruction, CFP positional tokens are used as Queries,
while OCT encoded tokens are used as Keys and Values:
\begin{equation}
\mathrm{Attn}_{O\rightarrow C}
=
\mathrm{softmax}
\left(
\frac{Q_C K_O^\top}{\sqrt{d}}
\right)
V_O .
\end{equation}

In both directions, the target modality provides only positional query tokens
and does not provide target visible tokens to the cross-modal decoder.
Therefore, the decoder must use source-modality features to predict the
target modality. This design encourages cross-modal prediction while reducing
the tendency to rely on target-modality shortcuts.

\subsection{Two-Stage Pretraining}
\label{sec:twostage}

We adopt a two-stage schedule to stabilize optimization. In Stage I, from
epoch 0 to 400, only the three intra-modal reconstruction objectives are
optimized. CFP, OCT, and CFP-Seg tokens are randomly masked, encoded by the
shared encoder, and reconstructed by their corresponding intra-modal decoders.
This stage provides a basic initialization for the shared encoder and allows
the model to learn CFP appearance, OCT structure, and CFP-derived anatomical
layout.

Stage I is not intended to enforce CFP--OCT alignment. Since each modality is
reconstructed from its own encoded features, the model can solve the task
largely using within-modality information. Thus, Stage I learns useful basic
representations, but does not explicitly require CFP features to predict OCT
structure or OCT features to predict CFP appearance.

In Stage II, from epoch 401 to 800, the two source-constrained cross-modal
decoders are introduced. The model continues to optimize the three intra-modal
reconstruction losses and additionally optimizes CFP-to-OCT and OCT-to-CFP
reconstruction losses. This stage preserves within-modality reconstruction
while adding an explicit cross-modal prediction objective.

\subsubsection{Training Objectives}
\label{sec:loss}

The Stage~I objective is a weighted sum of per-modality masked reconstruction
losses:
\begin{equation}
\label{eq:intra_loss}
  \mathcal{L}_{\text{intra}} =
    \lambda_{C}\,\mathcal{L}_{\text{MSE}}(I_C, \hat{I}_C)
  + \lambda_{O}\,\mathcal{L}_{\text{MSE}}(I_O, \hat{I}_O)
  + \lambda_{S}\,\mathcal{L}_{\text{CE}}(I_S, \hat{I}_S),
\end{equation}
where $I_C$, $I_O$, $I_S$ denote CFP, OCT, and CFP-Seg patches;
$\hat{\cdot}$ denotes reconstructed counterparts; and
$\lambda_C, \lambda_O, \lambda_S$ balance modality contributions.
Losses are computed over the masked token index set $\mathcal{M}$
($N{=}|\mathcal{M}|$):
\begin{align}
  \mathcal{L}_{\text{MSE}}(x,\hat{x}) &=
    \frac{1}{N}\sum_{i \in \mathcal{M}}\|x_i - \hat{x}_i\|_2^2, \\
  \mathcal{L}_{\text{CE}}(y,\hat{y})  &=
    -\frac{1}{N}\sum_{i \in \mathcal{M}}\sum_{c=1}^{C}
    y_{i,c}\log\hat{y}_{i,c}.
\end{align}

The Stage~II cross-modal objective reconstructs each modality conditioned
on the encoded features of the other:
\begin{equation}
\label{eq:cross_loss}
  \mathcal{L}_{\text{cross}} =
    \lambda_{C{\to}O}\,\mathcal{L}_{\text{MSE}}(I_O, \hat{I}_{O \leftarrow C})
  + \lambda_{O{\to}C}\,\mathcal{L}_{\text{MSE}}(I_C, \hat{I}_{C \leftarrow O}),
\end{equation}
where $\hat{I}_{O \leftarrow C}$ denotes OCT patches reconstructed from
CFP features via the source-constrained decoder. The
overall training objective is:
\begin{equation}
\label{eq:total_loss}
  \mathcal{L} =
    \lambda_{\text{intra}}\,\mathcal{L}_{\text{intra}}
  + \lambda_{\text{cross}}\,\mathcal{L}_{\text{cross}}.
\end{equation}

\subsection{Downstream Adaptation}
\label{sec:adaptation}

After pretraining, only the CFP branch is used for downstream evaluation. The
OCT input, CFP-Seg input, and all reconstruction decoders are discarded. For
classification, the pretrained CFP encoder is frozen and a linear classifier
is trained. For dense segmentation, a lightweight decoder is attached to the
frozen CFP encoder. For few-shot evaluation, the same protocol is repeated
with limited labeled examples per class. For feature-space visualization, CFP
embeddings are extracted from the frozen encoder and projected to two
dimensions using t-SNE for qualitative inspection of representation quality.

\section{Experiments and Results}
\label{sec:results}

\subsection{Implementation Details}
\label{sec:impl}

The shared encoder is a ViT-L model with embedding dimension $d{=}1024$, 24
transformer layers, and 16 attention heads. It is initialized from ImageNet
MAE weights \cite{deng2009imagenet,he2022masked}. Pretraining uses AdamW
\cite{loshchilov2017decoupled} with $\beta_1{=}0.9$, $\beta_2{=}0.95$,
weight decay $0.05$, cosine learning rate decay from $10^{-4}$ to $10^{-6}$,
and a 40-epoch linear warmup. The batch size is 256. Data augmentation
includes random horizontal flipping and random rotation.

Loss weights are set to $\lambda_C{=}\lambda_O{=}1.0$ and
$\lambda_S{=}0.5$ for intra-modal reconstruction. In Stage II,
$\lambda_{C\rightarrow O}{=}\lambda_{O\rightarrow C}{=}0.5$ is used for the
two cross-modal reconstruction losses. Pretraining is performed for 800
epochs, with Stage I occupying the first 400 epochs and Stage II the remaining
400 epochs.

\subsection{Downstream Evaluation Datasets}
\label{sec:datasets}

The downstream benchmark comprises 15 dataset-level evaluation settings across
two categories: CFP-based classification and dense segmentation. All
downstream evaluation cases were excluded from the pretraining
pool.

For CFP-based classification, we evaluated EyeMVP on 11 dataset-level settings.
These included common retinal and optic nerve disease classification tasks:
Messidor-2 for diabetic retinopathy~\cite{abramoff2013automated}, IDRID for
proliferative diabetic retinopathy~\cite{porwal2018indian}, IChallenge-AMD for
age-related macular degeneration~\cite{dt4f-rt59-20}, ORIGA-650 and REFUGE for
glaucoma~\cite{ORIGA,orlando2020refuge}, PALM-2 for pathological
myopia~\cite{55pk-8z03-19}, ODIR5K for multi-disease recognition~\cite{odir},
and a private retinal vein occlusion cohort. We also evaluated three
macular-disease cohorts that are difficult to assess from CFP alone: private
OCT-confirmed datasets for macular edema, myopic macular schisis, and central
serous retinopathy.

For dense segmentation, we used four CFP-based datasets: REFUGE for optic
disc/cup segmentation~\cite{orlando2020refuge}, DRIVE for retinal vessel
segmentation~\cite{1282003}, IDRID for hemorrhage segmentation
~\cite{porwal2018indian}, and e\_opthaEX for hard exudate segmentation
~\cite{decenciere2013teleophta}.

All private classification labels were assigned by two independent senior
ophthalmologists, using OCT as the reference standard when applicable.
Disagreements were resolved by a third adjudicating ophthalmologist. A concise summary of the downstream evaluation datasets is provided in
Table~\ref{tab:downstream_datasets}; detailed device information, image
resolutions, and label distributions are provided in Supplementary
Table~S6.

\begin{table}[t]
\caption{Summary of downstream evaluation datasets. The benchmark includes
11 classification settings and four segmentation settings. Representative
image examples are shown in Supplementary Figs.~S2 and~S3.}
\label{tab:downstream_datasets}
\centering
\renewcommand{\arraystretch}{0.96}
\setlength{\tabcolsep}{2.5pt}
\scriptsize
\begin{tabular}{lllc}
\toprule
Task & Dataset & Source & Samples \\
\midrule
DR cls & Messidor-2~\cite{abramoff2013automated} & Public & 1,748 \\
PDR cls & IDRID~\cite{porwal2018indian} & Public & 516 \\
AMD cls & IChallenge-AMD~\cite{dt4f-rt59-20} & Public & 400 \\
Glaucoma cls & ORIGA-650~\cite{ORIGA} & Public & 650 \\
Glaucoma cls & REFUGE~\cite{orlando2020refuge} & Public & 800 \\
PM cls & PALM-2~\cite{55pk-8z03-19} & Public & 800 \\
RVO cls & Private & Private & 1,067 \\
CSR cls & Private & Private & 590 \\
Multi-disease cls & ODIR5K~\cite{odir} & Public & 10,000 \\
ME cls & Private & Private & 600 \\
MS cls & Private & Private & 300 \\
\midrule
Disc/cup seg & REFUGE~\cite{orlando2020refuge} & Public & 1,200 \\
Vessel seg & DRIVE~\cite{1282003} & Public & 40 \\
HE seg & IDRID~\cite{porwal2018indian} & Public & 80 \\
EX seg & e\_opthaEX~\cite{decenciere2013teleophta} & Public & 48 \\
\bottomrule
\multicolumn{4}{l}{\footnotesize DR: diabetic retinopathy; PDR: proliferative diabetic retinopathy;} \\
\multicolumn{4}{l}{\footnotesize AMD: age-related macular degeneration; PM: pathological myopia;} \\
\multicolumn{4}{l}{\footnotesize RVO: retinal vein occlusion; CSR: central serous retinopathy;} \\
\multicolumn{4}{l}{\footnotesize ME: macular edema; MS: myopic macular schisis;} \\
\multicolumn{4}{l}{\footnotesize HE: hemorrhage; EX: hard exudate.}
\end{tabular}
\end{table}

\subsection{Evaluation Protocols}
\label{sec:protocols}

After pretraining, only the CFP branch was used for downstream evaluation.
For the private downstream
classification datasets, each patient contributed only one study eye, and
patients included in these datasets were excluded from the pretraining pool.
For full-data classification, the pretrained CFP encoder was frozen and a
linear classifier was trained on the extracted CFP features. Results were
evaluated using five-fold cross-validation and reported as mean $\pm$ SD
across folds. AUROC, balanced accuracy (BAcc), and average precision (AP)
were reported. For few-shot evaluation, $K\in\{4,8,16\}$ labeled samples per
class were sampled, and results were averaged over five random seeds.

For dense segmentation, the pretrained CFP encoder was frozen and a
ConvNeXt-based decoder \cite{liu2022convnet} was trained to predict
pixel-wise masks. Full-data segmentation results were also evaluated using
five-fold cross-validation and reported as mean $\pm$ SD across folds. Dice
coefficient and intersection-over-union (IoU) were reported. For feature-space
visualization, CFP embeddings were extracted from each frozen pretrained
encoder and projected to two dimensions using t-SNE~\cite{van2008visualizing}
with identical hyperparameters across all models; class labels were used only
to color the projected points and were not used during projection.





\subsection{Classification Performance}
\label{sec:cls_results}

Table~\ref{tab:cls_auroc_full_fewshot} summarizes the full-data and few-shot
classification AUROC across 11 dataset-level evaluation settings, covering
common retinal diseases, optic nerve disease, and macular conditions that are
difficult to identify from CFP alone. Under full-data linear probing, EyeMVP attains the highest or comparable
AUROC on most datasets relative to RETFound~\cite{zhou2023foundation},
VisionFM~\cite{qiu2024development}, and EyeCLIP~\cite{shi2025multimodal},
with the clearest gains on structurally demanding macular and optic-nerve
conditions. Against the strongest baseline, the largest improvements are on
ORIGA-650 glaucoma ($78.06\%$ vs.\ $72.26\%$) and myopic macular schisis
($86.67\%$ vs.\ $82.00\%$), followed by AMD ($94.81\%$ vs.\ $92.01\%$) and
macular edema ($92.26\%$ vs.\ $90.89\%$); on near-saturated PALM pathological
myopia it is on par with the best CFP baselines ($99.86\%$ vs.\ $99.92\%$).


A consistent pattern holds in the few-shot setting ($K{=}4$, $8$, $16$):
EyeMVP's largest gains again fall on the macular and optic-nerve tasks where
CFP alone is least informative, while it stays competitive elsewhere. At
$K{=}16$ it reaches $81.71\%$ AUROC on macular edema and $83.88\%$ on myopic
macular schisis ($6.69$ and $4.86$ points above the strongest baseline) and
$87.50\%$ on REFUGE glaucoma, whereas on near-saturated pathological myopia
the strong CFP baselines stay marginally ahead at low shot counts. This
indicates that the cross-modal pretraining improves data efficiency, with
benefits concentrated on structurally demanding conditions.


\begin{table*}[t]
\caption{Full-data and few-shot classification AUROC (\%).
Full-data results are from linear probing with the full training set.
Few-shot results are reported at $K{=}4,8,16$ labeled samples per class.
Values are mean $\pm$ SD. \textbf{Bold}: best performance. AP and BAcc are
reported in Supplementary Tables~S3 and~S4.}
\label{tab:cls_auroc_full_fewshot}
\centering
\renewcommand{\arraystretch}{0.96}
\setlength{\tabcolsep}{4.2pt}
\scriptsize
\begin{tabular}{ll l c c c c}
\toprule
Task & Dataset & Setting & RETFound & VisionFM & EyeCLIP & \textbf{EyeMVP} \\
\midrule
\multirow{4}{*}{DR cls}
& \multirow{4}{*}{Messidor-2~\cite{abramoff2013automated}}
& Full & 91.21 $\pm$ 2.20 & 87.35 $\pm$ 3.71 & 80.62 $\pm$ 2.99 & \textbf{92.69} $\pm$ 2.02 \\
& & $K{=}4$  & 73.32 $\pm$ 7.10 & 69.92 $\pm$ 4.73 & 60.79 $\pm$ 3.08 & \textbf{75.52} $\pm$ 3.71 \\
& & $K{=}8$  & 77.46 $\pm$ 5.95 & 76.97 $\pm$ 3.46 & 57.23 $\pm$ 7.01 & \textbf{77.71} $\pm$ 4.06 \\
& & $K{=}16$ & 78.37 $\pm$ 6.35 & 80.34 $\pm$ 1.88 & 58.58 $\pm$ 5.12 & \textbf{83.86} $\pm$ 1.86 \\
\midrule

\multirow{4}{*}{PDR cls}
& \multirow{4}{*}{IDRID~\cite{porwal2018indian}}
& Full & 82.21 $\pm$ 4.31 & 81.36 $\pm$ 4.02 & 68.91 $\pm$ 6.80 & \textbf{82.82} $\pm$ 3.32 \\
& & $K{=}4$  & 65.58 $\pm$ 7.98 & 65.57 $\pm$ 5.17 & 56.99 $\pm$ 8.18 & \textbf{75.32} $\pm$ 2.75 \\
& & $K{=}8$  & 75.30 $\pm$ 8.42 & 66.72 $\pm$ 7.87 & 57.37 $\pm$ 9.10 & \textbf{78.25} $\pm$ 5.93 \\
& & $K{=}16$ & 77.72 $\pm$ 5.72 & 68.59 $\pm$ 7.83 & 61.16 $\pm$ 4.59 & \textbf{79.75} $\pm$ 4.21 \\
\midrule

\multirow{4}{*}{AMD cls}
& \multirow{4}{*}{IChallenge-AMD~\cite{dt4f-rt59-20}}
& Full & 92.01 $\pm$ 2.74 & 91.76 $\pm$ 3.81 & 82.61 $\pm$ 7.97 & \textbf{94.81} $\pm$ 2.34 \\
& & $K{=}4$  & 79.32 $\pm$ 5.96 & 83.78 $\pm$ 3.50 & 70.88 $\pm$ 5.26 & \textbf{84.87} $\pm$ 3.31 \\
& & $K{=}8$  & 81.66 $\pm$ 8.94 & 85.01 $\pm$ 4.17 & 74.32 $\pm$ 4.83 & \textbf{86.22} $\pm$ 2.28 \\
& & $K{=}16$ & 87.86 $\pm$ 7.85 & \textbf{91.78} $\pm$ 2.30 & 74.45 $\pm$ 5.24 & 91.67 $\pm$ 1.47 \\
\midrule

\multirow{8}{*}{Glaucoma cls}
& \multirow{4}{*}{ORIGA-650~\cite{ORIGA}}
& Full & 72.26 $\pm$ 3.52 & 70.98 $\pm$ 3.38 & 68.84 $\pm$ 1.80 & \textbf{78.06} $\pm$ 2.84 \\
& & $K{=}4$  & 59.22 $\pm$ 3.65 & 63.89 $\pm$ 4.35 & 51.53 $\pm$ 3.11 & \textbf{66.23} $\pm$ 3.98 \\
& & $K{=}8$  & 62.68 $\pm$ 6.15 & 63.90 $\pm$ 5.74 & 53.99 $\pm$ 3.44 & \textbf{68.49} $\pm$ 3.89 \\
& & $K{=}16$ & 62.00 $\pm$ 4.43 & 60.34 $\pm$ 4.12 & 59.33 $\pm$ 5.60 & \textbf{70.13} $\pm$ 3.18 \\
\cmidrule{2-7}
& \multirow{4}{*}{REFUGE~\cite{orlando2020refuge}}
& Full & 95.87 $\pm$ 1.28 & 94.46 $\pm$ 2.28 & 85.44 $\pm$ 5.80 & \textbf{96.32} $\pm$ 2.15 \\
& & $K{=}4$  & 73.55 $\pm$ 5.12 & 75.65 $\pm$ 9.15 & 50.16 $\pm$ 7.21 & \textbf{76.70} $\pm$ 5.83 \\
& & $K{=}8$  & 81.51 $\pm$ 8.34 & 86.11 $\pm$ 7.40 & 58.78 $\pm$ 8.58 & \textbf{86.80} $\pm$ 4.35 \\
& & $K{=}16$ & 83.92 $\pm$ 4.80 & 86.83 $\pm$ 4.49 & 68.25 $\pm$ 2.45 & \textbf{87.50} $\pm$ 5.34 \\
\midrule

\multirow{4}{*}{PM cls}
& \multirow{4}{*}{PALM-2~\cite{55pk-8z03-19}}
& Full & 99.90 $\pm$ 0.07 & \textbf{99.92} $\pm$ 0.05 & 99.16 $\pm$ 0.60 & 99.86 $\pm$ 0.12 \\
& & $K{=}4$  & \textbf{99.95} $\pm$ 0.07 & 99.62 $\pm$ 0.12 & 87.34 $\pm$ 2.51 & 99.29 $\pm$ 0.23 \\
& & $K{=}8$  & \textbf{99.98} $\pm$ 0.02 & 99.79 $\pm$ 0.11 & 92.29 $\pm$ 2.88 & 99.41 $\pm$ 0.25 \\
& & $K{=}16$ & \textbf{99.97} $\pm$ 0.03 & 99.75 $\pm$ 0.13 & 94.15 $\pm$ 2.50 & 99.57 $\pm$ 0.19 \\
\midrule

\multirow{4}{*}{RVO cls}
& \multirow{4}{*}{Private}
& Full & 91.55 $\pm$ 2.30 & 91.81 $\pm$ 2.07 & 79.21 $\pm$ 1.53 & \textbf{92.92} $\pm$ 1.78 \\
& & $K{=}4$  & 60.06 $\pm$ 6.99 & \textbf{64.18} $\pm$ 7.16 & 47.45 $\pm$ 4.11 & 64.09 $\pm$ 5.02 \\
& & $K{=}8$  & 63.49 $\pm$ 7.66 & \textbf{74.33} $\pm$ 4.43 & 52.90 $\pm$ 4.00 & 74.10 $\pm$ 4.44 \\
& & $K{=}16$ & 68.98 $\pm$ 5.48 & 77.89 $\pm$ 7.02 & 52.01 $\pm$ 3.08 & \textbf{78.87} $\pm$ 4.09 \\
\midrule

\multirow{4}{*}{CSR cls}
& \multirow{4}{*}{Private}
& Full & 90.31 $\pm$ 2.92 & 85.70 $\pm$ 4.45 & 74.95 $\pm$ 3.88 & \textbf{91.49} $\pm$ 3.20 \\
& & $K{=}4$  & 53.56 $\pm$ 8.61 & 61.95 $\pm$ 6.26 & 48.01 $\pm$ 8.51 & \textbf{64.49} $\pm$ 4.23 \\
& & $K{=}8$  & 57.02 $\pm$ 10.30 & 67.08 $\pm$ 8.54 & 48.98 $\pm$ 4.64 & \textbf{67.91} $\pm$ 5.92 \\
& & $K{=}16$ & 64.90 $\pm$ 6.59 & 71.21 $\pm$ 3.35 & 52.96 $\pm$ 6.93 & \textbf{73.45} $\pm$ 3.78 \\
\midrule

\multirow{4}{*}{Multi-disease cls}
& \multirow{4}{*}{ODIR5K~\cite{odir}}
& Full & 77.28 $\pm$ 0.68 & 75.13 $\pm$ 0.77 & 69.21 $\pm$ 0.59 & \textbf{77.37} $\pm$ 0.99 \\
& & $K{=}4$  & 60.25 $\pm$ 1.71 & 64.12 $\pm$ 2.60 & 53.16 $\pm$ 1.17 & \textbf{65.37} $\pm$ 2.90 \\
& & $K{=}8$  & 62.76 $\pm$ 1.06 & 66.70 $\pm$ 1.04 & 56.15 $\pm$ 2.16 & \textbf{67.06} $\pm$ 2.04 \\
& & $K{=}16$ & 66.56 $\pm$ 1.89 & 69.31 $\pm$ 1.49 & 58.54 $\pm$ 3.69 & \textbf{72.14} $\pm$ 1.62 \\
\midrule

\multirow{4}{*}{ME cls}
& \multirow{4}{*}{Private}
& Full & 90.89 $\pm$ 1.77 & 88.78 $\pm$ 2.65 & 73.95 $\pm$ 2.59 & \textbf{92.26} $\pm$ 2.88 \\
& & $K{=}4$  & 62.87 $\pm$ 10.22 & 63.13 $\pm$ 6.66 & 47.01 $\pm$ 12.29 & \textbf{68.57} $\pm$ 6.08 \\
& & $K{=}8$  & 68.64 $\pm$ 8.48 & 68.24 $\pm$ 5.20 & 48.90 $\pm$ 15.44 & \textbf{71.35} $\pm$ 4.85 \\
& & $K{=}16$ & 74.98 $\pm$ 4.55 & 75.02 $\pm$ 2.76 & 52.93 $\pm$ 12.11 & \textbf{81.71} $\pm$ 2.14 \\
\midrule

\multirow{4}{*}{MS cls}
& \multirow{4}{*}{Private}
& Full & 76.29 $\pm$ 2.91 & 82.00 $\pm$ 3.05 & 75.10 $\pm$ 7.79 & \textbf{86.67} $\pm$ 2.10 \\
& & $K{=}4$  & 64.19 $\pm$ 8.70 & 61.74 $\pm$ 16.26 & 48.45 $\pm$ 8.21 & \textbf{68.57} $\pm$ 5.14 \\
& & $K{=}8$  & 72.88 $\pm$ 2.28 & 71.81 $\pm$ 7.73 & 57.24 $\pm$ 12.53 & \textbf{74.10} $\pm$ 3.67 \\
& & $K{=}16$ & 79.02 $\pm$ 4.04 & 71.76 $\pm$ 5.32 & 60.43 $\pm$ 9.26 & \textbf{83.88} $\pm$ 3.11 \\
\bottomrule
\multicolumn{7}{l}{\footnotesize DR: diabetic retinopathy; PDR: proliferative diabetic retinopathy; AMD: age-related macular degeneration;} \\
\multicolumn{7}{l}{\footnotesize PM: pathological myopia; RVO: retinal vein occlusion; CSR: central serous retinopathy;} \\
\multicolumn{7}{l}{\footnotesize ME: macular edema; MS: myopic macular schisis.}
\end{tabular}
\end{table*}

\begin{table*}[t]
\caption{Full-data and few-shot segmentation performance.
Full-data results are obtained with a frozen encoder and a trainable ConvNeXt decoder.
Few-shot results are reported at $K{=}4,8,16$ labeled samples.
Dice and IoU are reported in percentage. \textbf{Bold}: best performance.
Qualitative few-shot examples are shown in Supplementary Fig.~S1.}
\label{tab:seg_full_fewshot}
\centering
\renewcommand{\arraystretch}{0.92}
\setlength{\tabcolsep}{4.0pt}
\scriptsize
\begin{tabular}{ll l c c c c}
\toprule
Dataset (Task) & Setting & Metric & RETFound & VisionFM & EyeCLIP & \textbf{EyeMVP} \\
\midrule

\multirow{8}{*}{REFUGE~\cite{orlando2020refuge} (Disc/Cup)}
& \multirow{2}{*}{Full}
& Dice & 88.44 $\pm$ 0.50 & 91.15 $\pm$ 0.42 & 87.13 $\pm$ 0.41 & \textbf{94.48 $\pm$ 0.36} \\
& & IoU  & 80.29 $\pm$ 0.62 & 84.37 $\pm$ 0.49 & 79.34 $\pm$ 0.49 & \textbf{85.82 $\pm$ 0.36} \\
\cmidrule{2-7}
& \multirow{2}{*}{$K{=}4$}
& Dice & 71.69 $\pm$ 4.12 & 74.36 $\pm$ 3.85 & 77.56 $\pm$ 3.24 & \textbf{84.33 $\pm$ 2.95} \\
& & IoU  & 62.80 $\pm$ 4.45 & 67.66 $\pm$ 4.10 & 60.27 $\pm$ 3.44 & \textbf{78.76 $\pm$ 3.20} \\
\cmidrule{2-7}
& \multirow{2}{*}{$K{=}8$}
& Dice & 79.59 $\pm$ 2.21 & 81.45 $\pm$ 1.95 & 78.37 $\pm$ 1.88 & \textbf{89.51 $\pm$ 1.15} \\
& & IoU  & 68.77 $\pm$ 2.45 & 71.02 $\pm$ 2.15 & 64.45 $\pm$ 2.05 & \textbf{82.23 $\pm$ 1.40} \\
\cmidrule{2-7}
& \multirow{2}{*}{$K{=}16$}
& Dice & 86.32 $\pm$ 1.45 & 90.31 $\pm$ 1.12 & 83.55 $\pm$ 1.52 & \textbf{91.71 $\pm$ 0.85} \\
& & IoU  & 77.24 $\pm$ 1.68 & 83.09 $\pm$ 1.34 & 73.92 $\pm$ 1.25 & \textbf{85.58 $\pm$ 0.95} \\
\midrule

\multirow{8}{*}{DRIVE~\cite{1282003} (Vessels)}
& \multirow{2}{*}{Full}
& Dice & 85.10 $\pm$ 0.45 & 87.66 $\pm$ 0.38 & 84.91 $\pm$ 0.40 & \textbf{91.90 $\pm$ 0.25} \\
& & IoU  & 76.80 $\pm$ 0.52 & 79.54 $\pm$ 0.45 & 76.55 $\pm$ 0.46 & \textbf{87.25 $\pm$ 0.32} \\
\cmidrule{2-7}
& \multirow{2}{*}{$K{=}4$}
& Dice & 78.90 $\pm$ 3.15 & 84.49 $\pm$ 2.82 & 79.48 $\pm$ 2.95 & \textbf{88.65 $\pm$ 2.25} \\
& & IoU  & 69.35 $\pm$ 3.40 & 75.35 $\pm$ 3.05 & 68.85 $\pm$ 3.15 & \textbf{83.74 $\pm$ 2.45} \\
\cmidrule{2-7}
& \multirow{2}{*}{$K{=}8$}
& Dice & 79.88 $\pm$ 2.65 & 84.71 $\pm$ 2.42 & 80.07 $\pm$ 2.50 & \textbf{89.12 $\pm$ 1.95} \\
& & IoU  & 70.42 $\pm$ 2.88 & 75.71 $\pm$ 2.65 & 70.87 $\pm$ 2.72 & \textbf{83.96 $\pm$ 2.12} \\
\cmidrule{2-7}
& \multirow{2}{*}{$K{=}16$}
& Dice & 81.05 $\pm$ 1.10 & 84.73 $\pm$ 0.95 & 80.62 $\pm$ 1.02 & \textbf{90.66 $\pm$ 0.65} \\
& & IoU  & 71.98 $\pm$ 1.25 & 75.75 $\pm$ 1.10 & 71.39 $\pm$ 1.18 & \textbf{86.63 $\pm$ 0.78} \\
\midrule

\multirow{8}{*}{IDRID~\cite{porwal2018indian} (HE)}
& \multirow{2}{*}{Full}
& Dice & 59.50 $\pm$ 1.25 & 61.18 $\pm$ 1.15 & 56.85 $\pm$ 1.30 & \textbf{70.86 $\pm$ 0.95} \\
& & IoU  & 53.55 $\pm$ 1.35 & 55.38 $\pm$ 1.22 & 52.72 $\pm$ 1.42 & \textbf{68.54 $\pm$ 1.10} \\
\cmidrule{2-7}
& \multirow{2}{*}{$K{=}4$}
& Dice & 49.90 $\pm$ 2.15 & 50.80 $\pm$ 2.45 & 49.50 $\pm$ 1.82 & \textbf{58.28 $\pm$ 1.45} \\
& & IoU  & 45.30 $\pm$ 2.41 & 46.12 $\pm$ 2.55 & 47.75 $\pm$ 1.65 & \textbf{52.85 $\pm$ 1.72} \\
\cmidrule{2-7}
& \multirow{2}{*}{$K{=}8$}
& Dice & 52.48 $\pm$ 1.85 & 53.78 $\pm$ 1.64 & 51.20 $\pm$ 1.75 & \textbf{63.09 $\pm$ 1.25} \\
& & IoU  & 47.62 $\pm$ 2.05 & 48.80 $\pm$ 1.85 & 48.10 $\pm$ 1.54 & \textbf{57.12 $\pm$ 1.45} \\
\cmidrule{2-7}
& \multirow{2}{*}{$K{=}16$}
& Dice & 55.96 $\pm$ 1.35 & 56.85 $\pm$ 1.20 & 53.75 $\pm$ 1.28 & \textbf{67.92 $\pm$ 0.85} \\
& & IoU  & 50.75 $\pm$ 1.52 & 51.55 $\pm$ 1.42 & 50.15 $\pm$ 1.48 & \textbf{61.40 $\pm$ 1.05} \\
\midrule

\multirow{8}{*}{e\_opthaEX~\cite{decenciere2013teleophta} (EX)}
& \multirow{2}{*}{Full}
& Dice & 63.90 $\pm$ 0.93 & 67.50 $\pm$ 0.92 & 60.10 $\pm$ 0.39 & \textbf{82.98 $\pm$ 0.75} \\
& & IoU  & 56.79 $\pm$ 0.98 & 62.05 $\pm$ 0.76 & 58.68 $\pm$ 0.41 & \textbf{76.20 $\pm$ 0.88} \\
\cmidrule{2-7}
& \multirow{2}{*}{$K{=}4$}
& Dice & 51.26 $\pm$ 2.52 & 53.83 $\pm$ 2.15 & 49.18 $\pm$ 2.65 & \textbf{63.67 $\pm$ 1.25} \\
& & IoU  & 46.55 $\pm$ 2.80 & 48.34 $\pm$ 2.40 & 48.38 $\pm$ 1.92 & \textbf{58.59 $\pm$ 1.55} \\
\cmidrule{2-7}
& \multirow{2}{*}{$K{=}8$}
& Dice & 54.75 $\pm$ 1.85 & 58.34 $\pm$ 1.65 & 53.21 $\pm$ 2.05 & \textbf{64.38 $\pm$ 0.85} \\
& & IoU  & 49.68 $\pm$ 2.10 & 54.38 $\pm$ 2.92 & 51.08 $\pm$ 1.30 & \textbf{59.54 $\pm$ 1.15} \\
\cmidrule{2-7}
& \multirow{2}{*}{$K{=}16$}
& Dice & 59.18 $\pm$ 1.20 & 62.65 $\pm$ 1.10 & 56.85 $\pm$ 1.55 & \textbf{67.44 $\pm$ 0.67} \\
& & IoU  & 53.65 $\pm$ 1.45 & 56.55 $\pm$ 1.30 & 54.86 $\pm$ 1.78 & \textbf{64.89 $\pm$ 0.45} \\
\bottomrule

\multicolumn{7}{l}{\footnotesize HE: hemorrhage; EX: hard exudate.}
\end{tabular}
\end{table*}

\subsection{Dense Segmentation Performance}
\label{sec:seg_results}

\begin{figure}[t]
\centering
\includegraphics[width=\columnwidth]{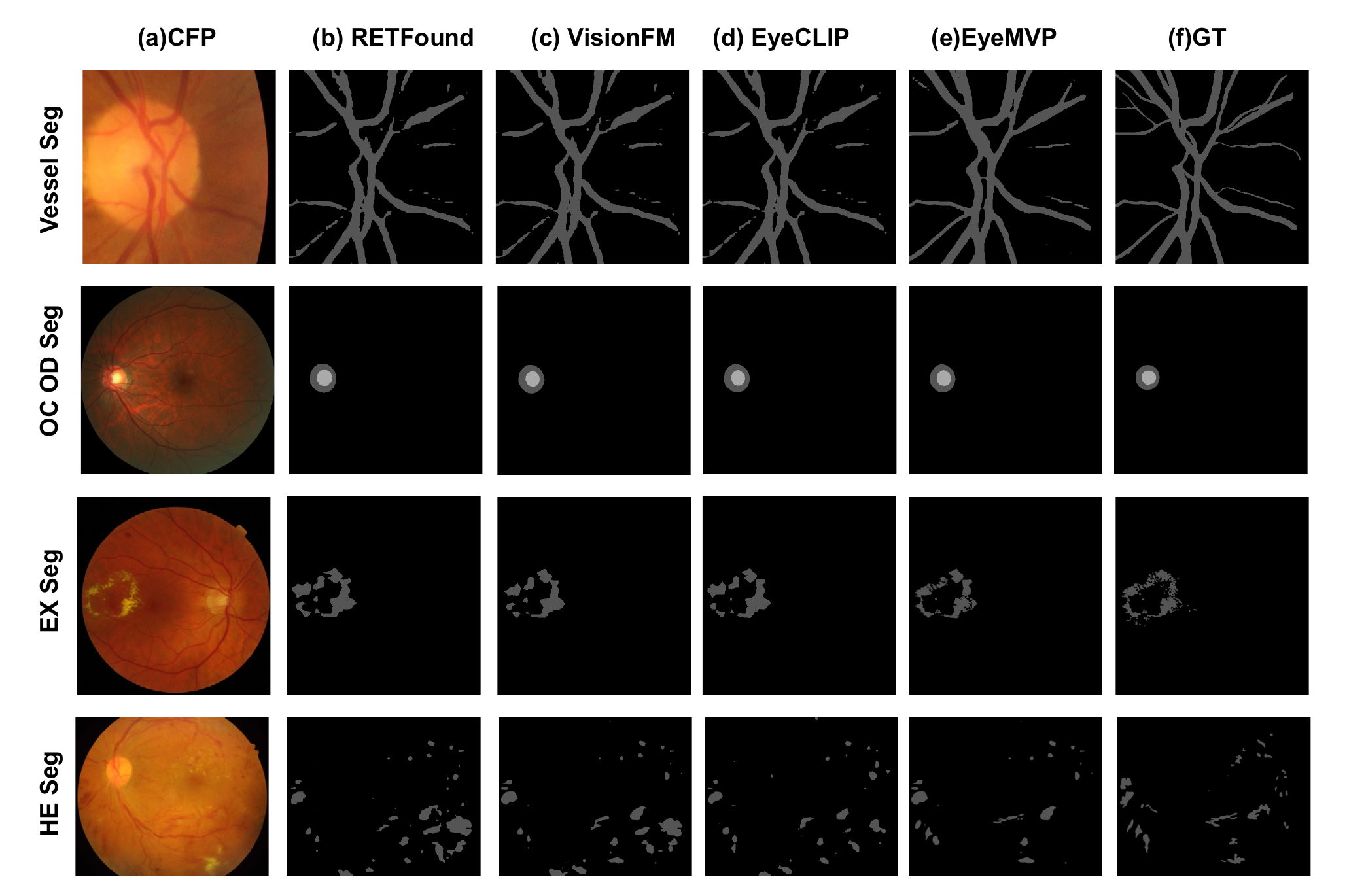}
\caption{Qualitative comparison of full-data segmentation results.
Columns show the input CFP image, predictions from RETFound,
VisionFM, EyeCLIP, EyeMVP, and the ground truth annotation. Rows correspond
to retinal vessel segmentation, optic disc/cup segmentation, hard exudate
segmentation, and hemorrhage segmentation. EyeMVP produces segmentation masks
that are generally closer to the ground truth, particularly for small lesion
regions and fine vascular structures.}
\label{fig:seg_visual}
\end{figure}

Table~\ref{tab:seg_full_fewshot} summarizes the full-data and few-shot
segmentation results on four CFP-based dense prediction tasks, including
optic disc/cup, retinal vessel, hemorrhage, and hard exudate segmentation.
All methods are evaluated using the same protocol, with a frozen encoder and
a trainable ConvNeXt decoder. In the full-data setting, EyeMVP achieves the
highest Dice and IoU on all four datasets. The improvements are more evident
on lesion segmentation tasks, where the target regions are small and often
have heterogeneous appearance. On IDRID hemorrhage segmentation, EyeMVP
achieves Dice/IoU of $70.86\%/68.54\%$, compared with
$61.18\%/55.38\%$ for VisionFM. On e\_opthaEX hard exudate segmentation,
EyeMVP achieves $82.98\%/76.20\%$, compared with $67.50\%/62.05\%$ for
VisionFM. Qualitative examples in Fig.~\ref{fig:seg_visual} show a similar
pattern: EyeMVP better preserves fine vascular structures and produces lesion
masks that are more consistent with the ground-truth annotations, whereas the
baseline models tend to miss small lesion regions or produce fragmented
predictions.

The few-shot segmentation results show a similar trend under limited
annotation. At $K{=}16$, EyeMVP obtains Dice/IoU of $91.71\%/85.58\%$ on
REFUGE disc/cup segmentation and $90.66\%/86.63\%$ on DRIVE vessel
segmentation. For lesion segmentation, EyeMVP reaches $67.92\%/61.40\%$ on
IDRID hemorrhage segmentation and $67.44\%/64.89\%$ on e\_opthaEX hard
exudate segmentation, outperforming the competing foundation models under
the same few-shot protocol. These results suggest that EyeMVP provides CFP
representations that are useful for dense prediction tasks, including
settings with limited pixel-level annotations.

\subsection{Feature-Space Visualization}
\label{sec:tsne}

To qualitatively assess the structure and class discriminability of the learned representation space, we visualized the CFP embeddings generated by frozen pretrained encoders using t-SNE~\cite{van2008visualizing}. For each model, CFP images were passed through the frozen encoder, and the resulting global feature vectors were obtained directly from the encoder output without task-specific fine-tuning. These features were then projected into a two-dimensional space using t-SNE with identical hyperparameters across all methods. No label information was used during projection; ground-truth labels were used only to color the projected points. We conducted this analysis on four representative tasks: diabetic retinopathy (DR) grading and three macular conditions that are difficult to assess from CFP alone, namely age-related macular degeneration (AMD), myopic macular schisis, and macular edema. RETFound~\cite{zhou2023foundation}, EyeCLIP~\cite{shi2025multimodal}, VisionFM~\cite{qiu2024development}, and EyeMVP were compared under the same embedding-extraction and projection protocol.

\begin{figure}[t]
\centering
\includegraphics[width=\columnwidth]{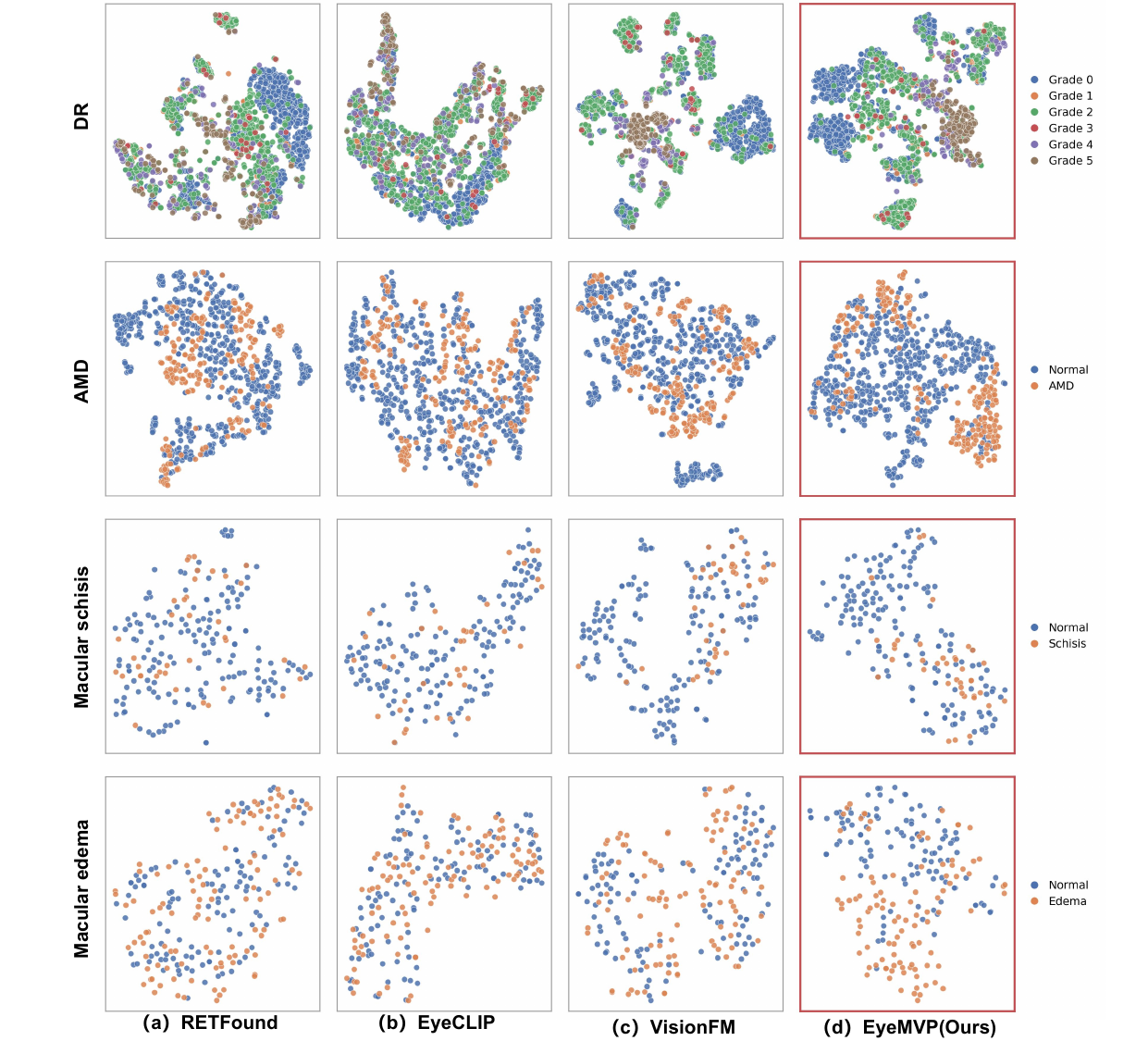}
\caption{t-SNE visualization of frozen CFP encoder embeddings for
RETFound, EyeCLIP, VisionFM, and EyeMVP (Ours). Each point corresponds to the
encoder output embedding of one CFP image projected to two dimensions; colors
denote ground-truth class labels and are not used during projection. Rows
correspond to four tasks: DR (six severity grades), AMD (normal
vs.\ AMD), myopic macular schisis (normal vs.\ schisis), and macular edema
(normal vs.\ edema). Compared with the baseline foundation models, EyeMVP
produces more class-coherent clusters with reduced overlap between disease and
normal populations, with the clearest improvements on the macular conditions.}
\label{fig:tsne}
\end{figure}

As shown in Fig.~\ref{fig:tsne}, EyeMVP produces embedding spaces with more coherent class structure than the baseline foundation models. For DR grading, the embeddings of EyeMVP form more clearly organized severity-related clusters. The non-referable population is concentrated in compact regions that are better separated from higher-severity grades, whereas the baseline embeddings show greater overlap between adjacent grades. This advantage is most evident for the macular conditions. For macular edema, EyeMVP produces substantially cleaner separation between normal and edema cases, whereas RETFound and EyeCLIP show considerable intermixing between the two classes. Similar trends are observed for AMD and myopic macular schisis, where the embeddings of EyeMVP form more contiguous disease regions with reduced overlap with the normal population.

These qualitative observations are consistent with the quantitative linear-probing results in Table~\ref{tab:cls_auroc_full_fewshot}. The macular conditions on which EyeMVP shows clear AUROC gains, including myopic macular schisis and macular edema, also show the most visibly improved class separation in the embedding space. Because the projection uses only the output of the frozen encoder without supervised adaptation, the improved cluster structure suggests that cross-modal CFP--OCT pretraining enables CFP representations to encode disease-relevant structural information in a more linearly separable form. This finding is consistent with the improved label efficiency observed in the few-shot experiments. We note that t-SNE is a nonlinear projection method intended for qualitative inspection, and absolute inter-cluster distances should not be interpreted quantitatively.

\subsection{Reader Study}
\label{sec:reader}

We conducted an exploratory reader study on an independent test set of 150
cases, including 50 myopic macular schisis, 50 macular edema, and 50 control
eyes. The test set was not used for pretraining, model selection, or
downstream fine-tuning. Nine ophthalmologists independently evaluated the CFP
images and were stratified by clinical experience into junior ($<$5 years),
intermediate (5--10 years), and senior ($>$10 years) groups, with three
readers in each group. Balanced accuracy (BAcc) was computed for each reader
and averaged within each experience group. EyeMVP was evaluated on the same
test set using CFP images only.

As shown in Table~\ref{tab:reader}, EyeMVP ($86.37\%$) exceeded the junior
($74.08\%$) and intermediate ($82.10\%$) groups but did not reach the
performance of senior ophthalmologists ($90.12\%$) on macular edema. For
myopic macular schisis, EyeMVP achieved a BAcc of $61.46\%$, compared with
$44.04\%$, $50.00\%$, and $55.36\%$ for the junior, intermediate, and senior
groups, respectively. These exploratory results suggest that EyeMVP may
provide complementary CFP-based information for macular conditions that are
difficult to assess from fundus photographs alone. Given the limited size of
the reader study, these findings should be interpreted as preliminary and
require validation in larger multi-reader cohorts.

\begin{table}[t]
\caption{Reader study on balanced accuracy (BAcc, \%) for myopic macular
schisis (MS) and macular edema (ME). Nine ophthalmologists were stratified
by experience level ($n{=}3$ per group) and compared with EyeMVP on the same
150-case independent test set.}
\label{tab:reader}
\centering
\renewcommand{\arraystretch}{0.98}
\setlength{\tabcolsep}{6pt}
\footnotesize
\begin{tabular}{lcc}
\toprule
Reader group & MS & ME \\
\midrule
Junior ($<$5 years)        & 44.04 & 74.08 \\
Intermediate (5--10 years) & 50.00 & 82.10 \\
Senior ($>$10 years)       & 55.36 & \textbf{90.12} \\
EyeMVP                     & \textbf{61.46} & 86.37 \\
\bottomrule
\end{tabular}
\end{table}

\subsection{Ablation Studies}
\label{sec:ablation}

Table~\ref{tab:ablation} reports component ablations on three tasks where
retinal or macular structure is expected to be important: ORIGA-650 glaucoma,
macular edema, and myopic macular schisis. Each ablation removes one
pretraining component while using the same CFP encoder architecture and
otherwise matched pretraining and downstream linear probing conditions. We
report representative ablations in the main text; complete results across
all tasks are provided in Supplementary Table~S5.

Removing the OCT modality from pretraining leads to the largest performance
decrease across the three tasks, with AUROC drops of $5.53$, $11.60$, and
$12.57$ percentage points on ORIGA-650, macular edema, and myopic macular
schisis, respectively. This result supports the role of paired CFP--OCT data
in providing useful cross-modal supervision for CFP representation learning.

The Stage-I-only variant retains the shared encoder and the three intra-modal
reconstruction decoders, but removes the source-constrained CFP-to-OCT and
OCT-to-CFP reconstruction objectives introduced in Stage II. This setting
corresponds to a shared-encoder multimodal masked autoencoding baseline,
where each modality is reconstructed primarily from its own visible tokens.
Compared with the full two-stage model, Stage-I-only pretraining reduces
AUROC by $4.33$, $7.90$, and $6.12$ percentage points on ORIGA-650, macular
edema, and myopic macular schisis, respectively. These results indicate that
explicit cross-modal reconstruction provides additional supervision beyond
shared-encoder intra-modal reconstruction.

\begin{figure}[t]
\centering
\includegraphics[width=\columnwidth]{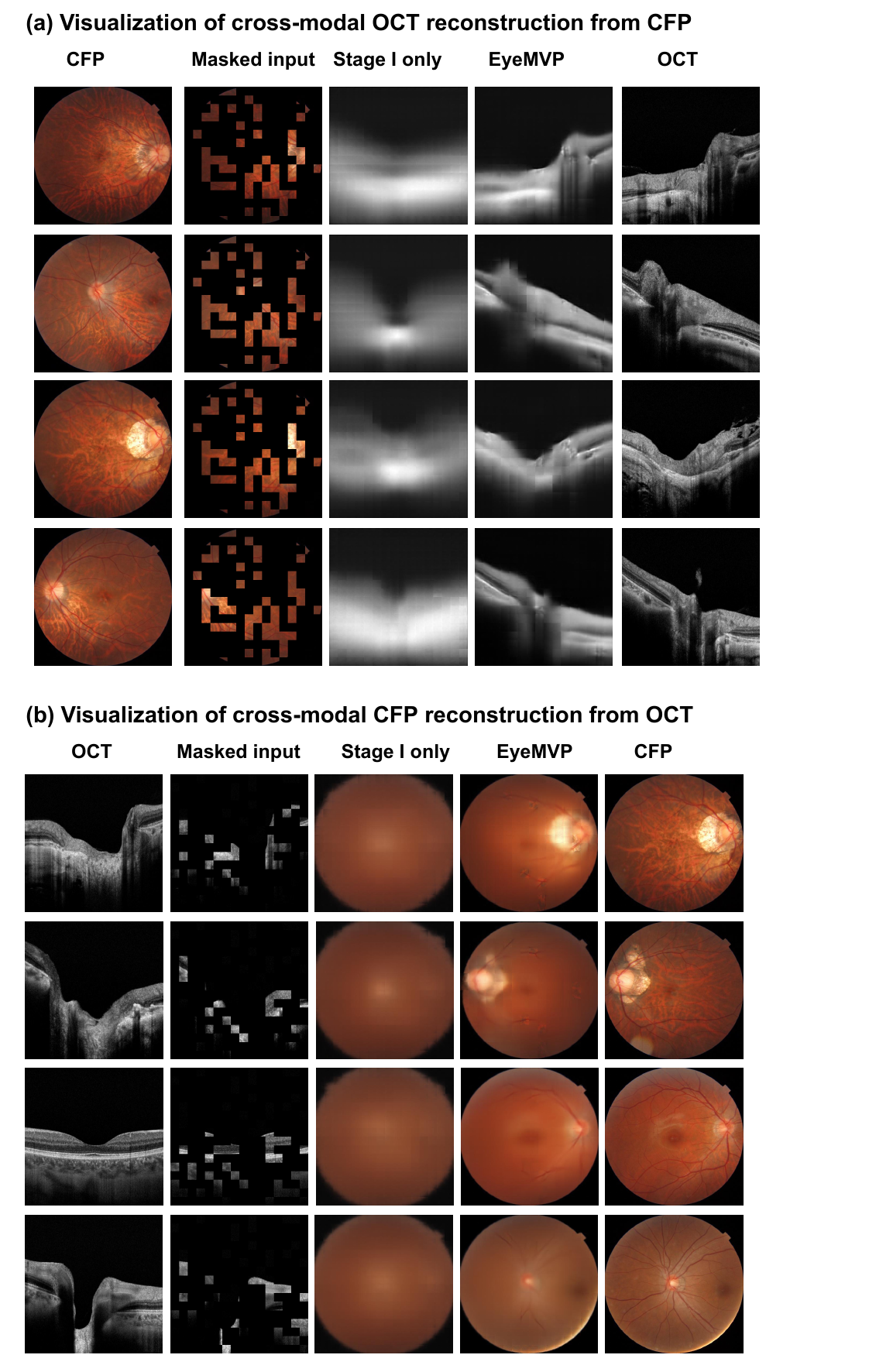}
\caption{Qualitative visualization of bidirectional cross-modal
reconstruction during pretraining. (a)~OCT reconstruction from partially
visible CFP tokens. Columns show the CFP image, masked CFP input, Stage-I-only
reconstruction, EyeMVP reconstruction with source-constrained cross-modal
decoding, and the paired OCT target. (b)~CFP reconstruction from partially
visible OCT tokens, shown in the reverse direction with the paired fundus
image as the target reference. Compared with the Stage-I-only variant, EyeMVP
recovers sharper modality-specific structures, including OCT-like retinal
contours and more fundus-like vascular and disc appearance. These
visualizations illustrate the cross-modal pretraining objective and are not
intended to represent diagnostic OCT or CFP synthesis.}
\label{fig:cross_modal_recon}
\end{figure}

To further examine this component, we visualize CFP-to-OCT reconstruction
under a source-constrained setting in Fig.~\ref{fig:cross_modal_recon}, where
all OCT tokens are masked and only partial CFP tokens are provided. Compared
with the Stage-I-only variant, EyeMVP produces reconstructions with clearer
OCT-like retinal contours and layer-related contrast. These visualizations
illustrate the pretraining objective and are not intended to represent
diagnostic OCT synthesis.

Removing the CFP-Seg branch decreases AUROC by $4.91$ percentage points on
ORIGA-650 and $4.03$ percentage points on myopic macular schisis. This
suggests that CFP-derived anatomical masks provide useful spatial guidance
during pretraining by supplying coarse fundus-layout context for cross-modal
reconstruction. Overall, the ablation results show that the paired OCT
modality, Stage-II cross-modal reconstruction, and CFP-derived structural
masks each contribute to the final EyeMVP performance.

\begin{table}[t]
\caption{Component ablation AUROC (\%) on structurally relevant classification
tasks. Values in parentheses denote absolute AUROC drop ($\Delta$) relative
to the full model. Mean values reported; SD\,$\leq$\,2.91\% across all entries.}
\label{tab:ablation}
\centering
\renewcommand{\arraystretch}{1.05}
\setlength{\tabcolsep}{4pt}
\footnotesize
\begin{tabular}{lccc}
\toprule
Configuration & ORIGA-650 & ME & MS \\
\midrule
w/o OCT modality    & 72.53 ($-$5.53) & 80.66 ($-$11.60) & 74.10 ($-$12.57) \\
w/o Stage-II cross recon. & 73.73 ($-$4.33) & 84.36 ($-$7.90)  & 80.55 ($-$6.12)  \\
w/o CFP-Seg         & 73.15 ($-$4.91) & 85.23 ($-$7.03)  & 82.64 ($-$4.03)  \\
\midrule
\textbf{EyeMVP (full)} & \textbf{78.06} & \textbf{92.26} & \textbf{86.67} \\
\bottomrule
\end{tabular}
\end{table}

\section{Discussion}
\label{sec:discussion}

This study presents EyeMVP, a cross-modal retinal foundation model that uses
paired CFP--OCT data to improve CFP-based representation learning. Across
classification, segmentation, few-shot learning, feature-space visualization,
and reader-study evaluations, EyeMVP shows consistent gains over representative
retinal foundation models. The improvements are most evident on tasks in
which retinal or macular structure is clinically relevant but not directly
resolved in CFP, such as glaucoma, macular edema, and myopic macular schisis.
These findings support the value of using OCT as a structural supervisory
signal during CFP pretraining, while retaining the practical advantage that
only CFP images are required at inference.

The design of EyeMVP reflects several practical challenges in CFP--OCT
pretraining. CFP and OCT are acquired in different imaging geometries, and
their image tokens do not share a common spatial grid. A purely intra-modal masked
autoencoding objective may therefore learn modality-specific reconstruction
shortcuts without sufficiently encouraging cross-modal interaction. EyeMVP
addresses this issue using source-constrained cross-modal decoding, which
requires one modality to predict the other from source-modality
representations. The CFP-Seg branch provides coarse anatomical guidance for
the fundus image, and the Dirichlet masking strategy exposes the model to
variable modality availability during pretraining. Ablation results and
reconstruction visualizations suggest that these components contribute
complementary benefits. We use the two-stage schedule as an implementation
choice to stabilize optimization rather than as an independent methodological
contribution.

EyeMVP should not be interpreted as a replacement for OCT. The model does not
measure retinal depth or synthesize diagnostic OCT images; instead, it learns
CFP representations that are informed by statistical associations with paired
OCT data. This distinction is important for clinical deployment. In settings
where OCT is unavailable, such representations may help improve CFP-based
screening and triage, but positive or uncertain findings should still be
confirmed with standard clinical examination and OCT when indicated. The
exploratory reader study further suggests that EyeMVP may provide
complementary CFP-based information for macular conditions that are difficult to assess
from CFP alone, but larger multi-reader and prospective studies are required
before drawing clinical conclusions.

Despite its strong performance, we acknowledge the methodological limits of the
approach. EyeMVP learns statistical CFP--OCT structural correspondence during
cross-modal pretraining rather than a registered or physically grounded
mapping, relying on fine-grained texture and vascular patterns as learned
surrogates for retinal structure. Because the CFP--OCT pairs are linked at the
eye-visit rather than the pixel level, it captures population-level
correspondence rather than exact spatial alignment, a trade-off that enables
large-scale pretraining while bounding spatial precision. As these learned
correlations are bounded by the pretraining distribution, performance may
degrade when the underlying cues are disrupted, such as under media opacities
(e.g., dense cataracts or vitreous hemorrhage) or in rare phenotypes and
post-surgical eyes that violate normal CFP--OCT relationships. Future
deployments should incorporate uncertainty quantification to flag such
out-of-distribution cases for clinical review.


\section{Conclusion}
\label{sec:conclusion}

We presented EyeMVP, a cross-modal retinal foundation model that exploits
paired CFP--OCT data during pretraining to enrich CFP representations while
requiring only CFP images at inference. Its core novelty is a
source-constrained cross-modal reconstruction scheme that forces CFP features
to predict OCT structure without access to target-modality tokens,
complemented by an auxiliary CFP-Seg branch that injects anatomical-layout
priors. Across classification, dense segmentation, and few-shot settings,
EyeMVP is competitive with or better than representative retinal foundation
models, with the clearest gains on glaucoma and macular conditions, where
structural cues are clinically important but poorly resolved in CFP. By
transferring OCT-derived structural information into a CFP-only model, EyeMVP
offers a practical path to structure-aware fundus analysis where OCT is
unavailable.

\section*{Declarations}
\subsection{Data and Code availability}
The pre-training data was collected in-house and is not publicly available due to privacy restrictions. Downstream evaluation was conducted on 11 public datasets and 4 private in-house datasets. The public datasets can be accessed via the following links: Messidor-2 \cite{abramoff2013automated} (\url{https://www.adcis.net/en/third-party/messidor2/}), IDRID \cite{porwal2018indian} (\url{http://idrid.grand-challenge.org/grading}), IChallenge-AMD \cite{dt4f-rt59-20} (\url{https://amd.grand-challenge.org/}), ORIGA-650 \cite{ORIGA} (\url{https://doi.org/10.57702/hoijctum}), REFUGE \cite{orlando2020refuge} (\url{https://refuge.grand-challenge.org/Home2020/grading}), PALM \cite{55pk-8z03-19} (\url{https://palm.grand-challenge.org/}), ODIR5K \cite{odir} (\url{https://odir2019.grand-challenge.org/introduction/}), DRIVE \cite{1282003} (\url{https://drive.grand-challenge.org/}), REFUGE \cite{orlando2020refuge} (\url{https://refuge.grand-challenge.org/Home2020/segmentation}), e\_opthaEX \cite{decenciere2013teleophta} (\url{https://www.adcis.net/en/third-party/e-ophtha/}), and IDRID \cite{porwal2018indian} (\url{http://idrid.grand-challenge.org/segmentation}). While the 4 in-house evaluation datasets cannot be fully shared due to patient privacy regulations. The dataset, code, and models will be publicly available at \url{https://github.com/ML-AILab/EyeMVP}.

\section*{references}
\bibliographystyle{IEEEtran}
\bibliography{references}





\end{document}